\documentclass[conference]{IEEEtran}
\usepackage{times}
\usepackage{graphicx} 
\usepackage{subfigure} 
\usepackage{algorithm}
\usepackage{algorithmic}
\usepackage{amsmath}
\usepackage{bm}
\usepackage{widetable}
\usepackage{multirow}
\usepackage{amssymb}

\graphicspath{{image_pdf/}}

\newcommand{\linebreakcell}[2][c]{%
  \begin{tabular}[#1]{@{}c@{}}#2\end{tabular}}

\ifCLASSINFOpdf
\else
\fi
\begin{document}
%
\title{FPGA-based Low-power Speech Recognition\\with Recurrent Neural Networks}

\author{\IEEEauthorblockN{Minjae Lee, Kyuyeon Hwang, Jinhwan Park, Sungwook Choi, Sungho Shin and Wonyong Sung}
\IEEEauthorblockA{Department of Electrical and Computer Engineering, Seoul National University\\
1, Gwanak-ro, Gwanak-gu, Seoul, 08826 Korea\\
\{mjlee, khwang, jhpark, swchoi, shshin\}@dsp.snu.ac.kr, wysung@snu.ac.kr}}


%


\maketitle

\begin{abstract}
In this paper, a neural network based real-time speech recognition (SR) system is developed using an FPGA for very low-power operation. The implemented system employs two recurrent neural networks (RNNs); one is a speech-to-character RNN for acoustic modeling (AM) and the other is for character-level language modeling (LM). The system also employs a statistical word-level LM to improve the recognition accuracy. The results of the AM, the character-level LM, and the word-level LM are combined using a fairly simple $N$-best search algorithm instead of the hidden Markov model (HMM) based network. The RNNs are implemented using massively parallel processing elements (PEs) for low latency and high throughput. The weights are quantized to 6 bits to store all of them in the on-chip memory of an FPGA. The proposed algorithm is implemented on a Xilinx XC7Z045, and the system can operate much faster than real-time. 
\end{abstract}


%
\IEEEpeerreviewmaketitle

\section{Introduction}
\label{sec_introduction}

Speech recognition has long been studied, and most of the algorithms employ hidden Markov models (HMMs) or its variants as inference and information combining tools \cite{huang2001spoken, mohri2002weighted}. Recently, deep neural networks are employed for acoustic modeling (AM) of state of the art speech recognition systems which, however, are not free from the HMM \cite{hinton2012deep}. HMM modeling for speech recognition demands a vast amount of memory access operations on a large size network, whose memory capacity usually exceeds a few hundred megabytes \cite{you2009parallel}. Thus, speech recognition algorithms are usually implemented on GPUs or multi-core systems that equip large DRAM-based memory, which are hardly power efficient. 

Recently, fully neural recurrent network based speech recognition algorithms are actively investigated \cite{maas2015lexicon, hwang2016character}. 
The RNN is end-to-end trained with connectionist temporal classification (CTC) \cite{graves2006ctc} to directly transcribe the input utterance to characters. The RNN has also been used for language modeling (LM), which shows much better capability than tri-gram based statistical  algorithms \cite{sundermeyer2015feedforward}. Recently, complete speech recognition algorithms have been developed by combining the CTC RNN and the RNN LM \cite{maas2015lexicon, hwang2016character}. These RNN based algorithms do not employ a conventional HMM that needs a large search space. However, neural network algorithms, including RNNs, demand a very large number of arithmetic operations, thus they are mostly implemented using GPUs \cite{amodei2016deep, miao2015eesen}. 

In this work, a low-power real-time speech recognition (SR) system is developed using an FPGA. The developed system employs two long-short term memory (LSTM) RNNs \cite{hochreiter1997long}; one for acoustic modeling and the other for character-level language modeling. A statistical word-level LM is also used to further improve the recognition performance. The overall algorithm is shown in \figurename~\ref{fig_algorithmOV}. The information generated from the RNNs and the word-level LM is combined using a tree structured $N$-best beam search algorithm. The beam search employing the beam width of 128 only requires about 197 KB of data structure, while the conventional HMM based network demands a few hundred megabytes of memory. The SR system employs a unidirectional RNN based acoustic model, causing a slight disadvantage in the recognition performance  when compared to a bidirectional one, but is more appropriate for online real-time applications where immediate reaction to utterance is desired. 

The RNNs for acoustic modeling and character-level LM are implemented on a mid-sized FPGA, Xilinx XC7Z045, which contains 2.18 MB on-chip memory. To store all the weights of the RNNs in the on-chip memory, the weights are quantized to 6 bits using the retraining based fixed-point optimization algorithm \cite{hwang2014fixed}. The RNN for the character-level LM stores 128 contexts in the on-chip memory, where each context is assigned to each beam in the $N$-best search. All of the weights and the contexts are stored in the on-chip memory of the FPGA, and thus the RNNs do not need DRAM accesses which require a large amount of energy \cite{horowitz2014computings, han2016eie}. As a result, this speech recognition system only uses DRAM accesses for tri-gram based language modeling, and consumes very small power compared to GPU based systems or other off-chip memory based architectures. The RNNs in the FPGA are implemented using highly parallel arithmetic arrays.

The paper is organized as follows. In Section~\ref{sec_relatedWorks}, recent related works are revisited. Section~\ref{sec_SR} describes the implemented SR algorithm. The FPGA based implementation of the algorithm is shown in Section~\ref{sec_FPGA}. The system is evaluated in Section \ref{sec_exp}. Concluding remarks are in Section \ref{sec_conclusion}.

\section{Related works}
\label{sec_relatedWorks}

\subsection{Large Vocabulary Continuous Speech Recognition}
\label{subsec_prevLVCSR}

Most state-of-the-art large vocabulary continuous speech recognition (LVCSR) systems employ a DNN-HMM hybrid acoustic model \cite{hinton2012deep} or a weighted finite state transducer (WFST) decoder \cite{mohri2002weighted}. The WFST network is composed by integrating the HMM acoustic model, a pronunciation lexicon model, and a word-level n-gram back-off language model. Therefore, the resulting decoding network becomes huge, which is usually over a few hundred megabytes \cite{you2009parallel}, and hinders small-footprint low-power implementations.

A traditional LVCSR performs Viterbi decoding \cite{forney1973viterbi} on the WFST network using senone-level likelihoods computed by the acoustic model. Efficient hardware based implementation of the LVCSR \cite{choi2010fpga} is difficult because of the large amount of search operations needed for Viterbi decoding. Specifically, the network cannot be embedded in the on-chip memory due to its size and is usually stored on an off-chip DRAM module. The energy cost of a DRAM access is large since static power is required to keep the I/O active and data must travel a long distance \cite{horowitz2014computings}. Therefore, the decoding procedure on WFST using DRAM consumes a large amount of power. 

Recently, several RNN based end-to-end speech recognizers have been developed \cite{graves2014towards, amodei2016deep, miao2015eesen}. A phoneme-level CTC-trained RNN for acoustic modeling can reduce the size of a WFST network to about a half of that needed for DNN-HMM hybrid models \cite{miao2015eesen}. Also, character-level RNN language models and prefix beam search decoding greatly reduce the complexity of the decoding stage \cite{maas2015lexicon, hwang2016character}. Especially, a tree-based online decoding algorithm is proposed for low-latency speech recognition \cite{hwang2016character}.

\subsection{FPGA-Based Neural Network Implementation}
\label{subsec_prevFPGA}

Neural networks demand many multiply and add operations, but they are hardware-friendly in nature due to their massive parallelism. However, many previous implementations store the network parameters on an external DRAM, since the networks usually demand more than millions of parameters. Note that the weights for fully connected layers or recurrent neural networks are used only once when fetched, thus their accesses show very low temporal locality. There have been efforts to reduce the size of parameters by quantization. The bit-width of DNNs can be reduced to only two bits by retraining the quantized parameters with a modified backpropagation algorithm \cite{hwang2014fixed}. This approach was successfully applied to CNNs and RNNs \cite{anwar2015fixed, shin2016fixed}. RNNs also demand a large number of parameters. Thus, it is helpful to quantize the parameters in low bits. A study on weight quantization of RNNs was presented in \cite{shin2016fixed}. The retrain-based quantization method led to an efficient VLSI implementation of DNNs that store all the quantized parameters on the on-chip SRAM \cite{kim2014x1000}. Also, a similar architecture was employed for a DNN implementation on an FPGA \cite{park2016fpga}. 

\section{Speech Recognition without HMM}
\label{sec_SR}

\subsection{Algorithm Overview}
\label{subsec_overview}

\begin{figure}[t]
\begin{center}
\centerline{\includegraphics[width=70mm]{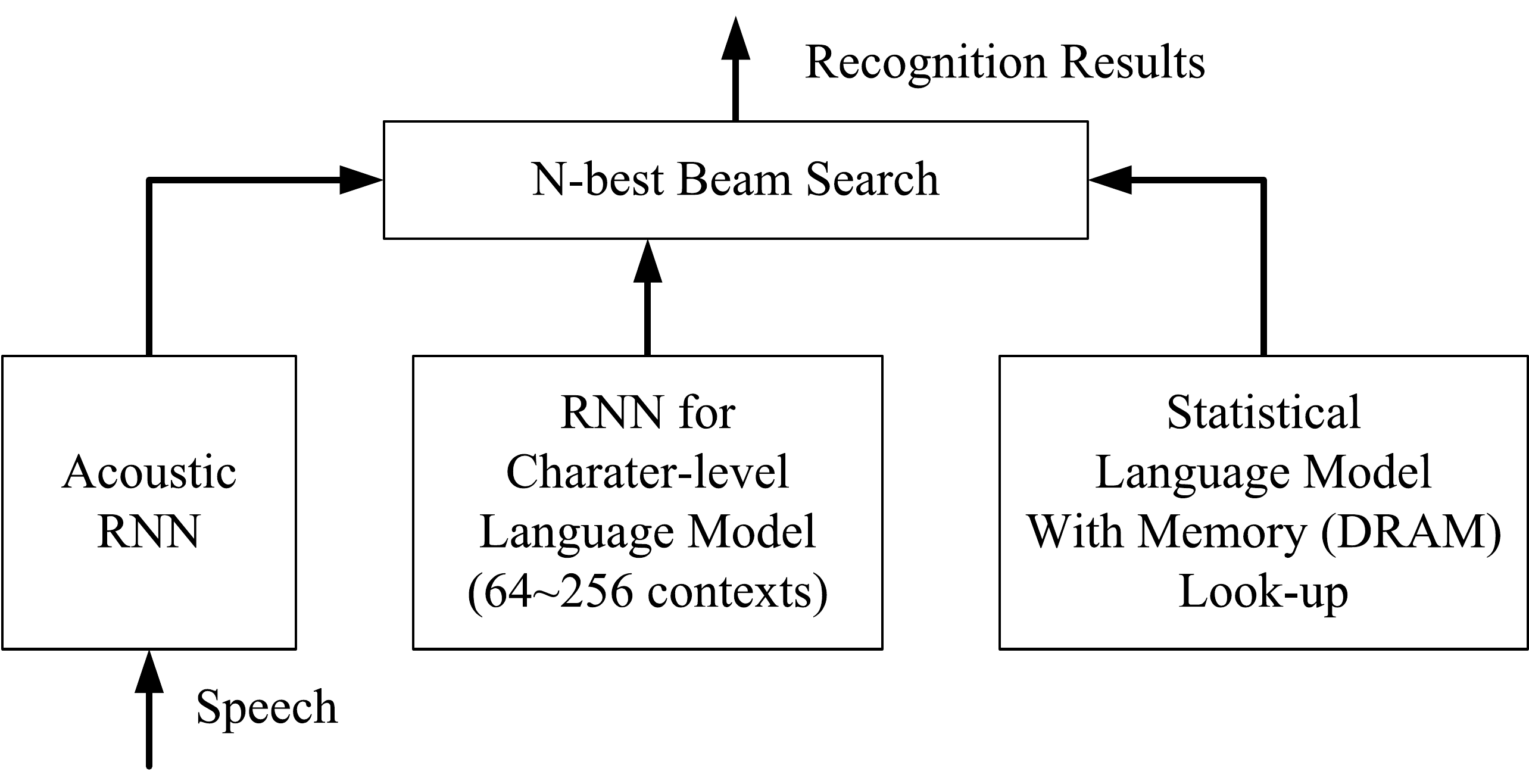}}
\vskip -0.15in
\caption{Structure of the proposed speech recognition system.}
\label{fig_algorithmOV}
\end{center}
\vskip -0.4in
\end{figure} 
The speech recognition algorithm implemented in this paper consists of an RNN for acoustic modeling (AM), an RNN for character-level LM and a statistical word-level LM as illustrated in \figurename~\ref{fig_algorithmOV}. The RNN AM employs the online CTC algorithm \cite{hwang2016sequence} and generates the probabilities of characters by analyzing each frame of input utterance. The character-level RNN LM outputs the probabilities of the following characters, while the statistical word-level tri-gram back-off LM shows that of the following words. The information generated from these three modules are integrated to find the best hypothesis using an $N$-best search algorithm.  

The acoustic model has a deep LSTM network structure and is end-to-end trained with online CTC algorithm \cite{hwang2016sequence}. Although some recent RNN-based end-to-end speech recognition algorithms \cite{graves2014towards, amodei2016deep, miao2015eesen} employ the bidirectional structure for recognition performance improvement, we use a unidirectional structure for real-time operation, where it is not allowed to access the future contexts.

The proposed SR system also employs a deep unidirectional LSTM RNN for character-level LM \cite{sutskever2011generating}. Since the character-level LM does not utilize any lexicon information, it can dictate out of vocabulary (OOV) words but is slightly disadvantaged in recognizing vocabularies in the dictionary. When compared to widely used HMM or RNN based speech recognition algorithms, the implemented one has the capability of low-latency decoding and OOV dictation, but these characteristics also mean slight weakness in the recognition accuracy. The structures of the RNNs for the AM and character-level LM are described in \cite{hwang2016character}. 

In our work, conventional statistical tri-gram back-off model is also employed for the word-level LM to complement the RNN based character-level LM. For better backing-off, we use improved Kneser-Ney smoothing \cite{kneser1995improved}. The word-level LM is integrated for the $N$-best beam search in a similar manner as the character-level LM \cite{hwang2016character}, except that the rescoring is performed on the fly, only when the active node represents a blank or the end of sentence (EOS) symbol. Also, the word insertion bonus is considered when the word-level LM is applied. 
Note that the number of DRAM accesses for the word-level LM is not very large. 

\subsection{Beam Search Algorithm}
\label{subsec_beamSearch}

In this work, the beam search decoding is conducted with a simple prefix tree structure. The $N$-best hypotheses are generated using the RNN AM and the RNN for character-level LM, and rescored by the statistical word-level LM on the fly. 

Let $L$ be the set of all output labels in the RNN AM except for the CTC blank. The input feature vector from time 1 to $t$ is denoted as $\mathbf{x}_{1:t}$. Given $\mathbf{x}_{1:t}$, the goal of the beam search decoding is to find the label sequence with the maximum posterior probability generated by the RNN AM.

The hypotheses are represented by a simple tree, where each node in the tree represents labels in $L$. To deal with CTC state transitions, state-based networks that are represented with CTC states, $L'=L\,\cup\,$\{CTC blank\}, are employed in low level by decomposing a tree node into two CTC states; a state that corresponds to a label in $L′$ and a following state that represents the CTC blank label. 

Since the tree grows indefinitely as the beam search proceeds, it is necessary to prune the search tree periodically. The tree is pruned both in depth and width as explained in \cite{hwang2016character}.

\subsection{Retraining Based Fixed-Point Optimization}
\label{subsec_fixedpoint}

Since the LSTM RNN contains millions of weights, an FPGA based implementation demands large on-chip memory space to store the parameters. It is not efficient to store the weights on the external DRAM because the fetched weights are used only once for each output computation. In our implementation, the retraining based method \cite{hwang2014fixed, shin2016fixed} is applied to reduce the word-length of weights. The algorithm groups the weights and signals by layer, applies direct quantization to each group, and retrains the whole network in the quantized domain. In our work, the weights and the internal signals are quantized to 6 and 8 bits, respectively. We find that the internal LSTM cells demand high precision, and thus, they are represented in 16 bits.

\section{FPGA-Based Implementation}
\label{sec_FPGA}

\subsection{Overview of the FPGA System}
\label{subsec_overviewFPGA}

\begin{figure}[t]
\begin{center}
\centerline{\includegraphics[width=\columnwidth]{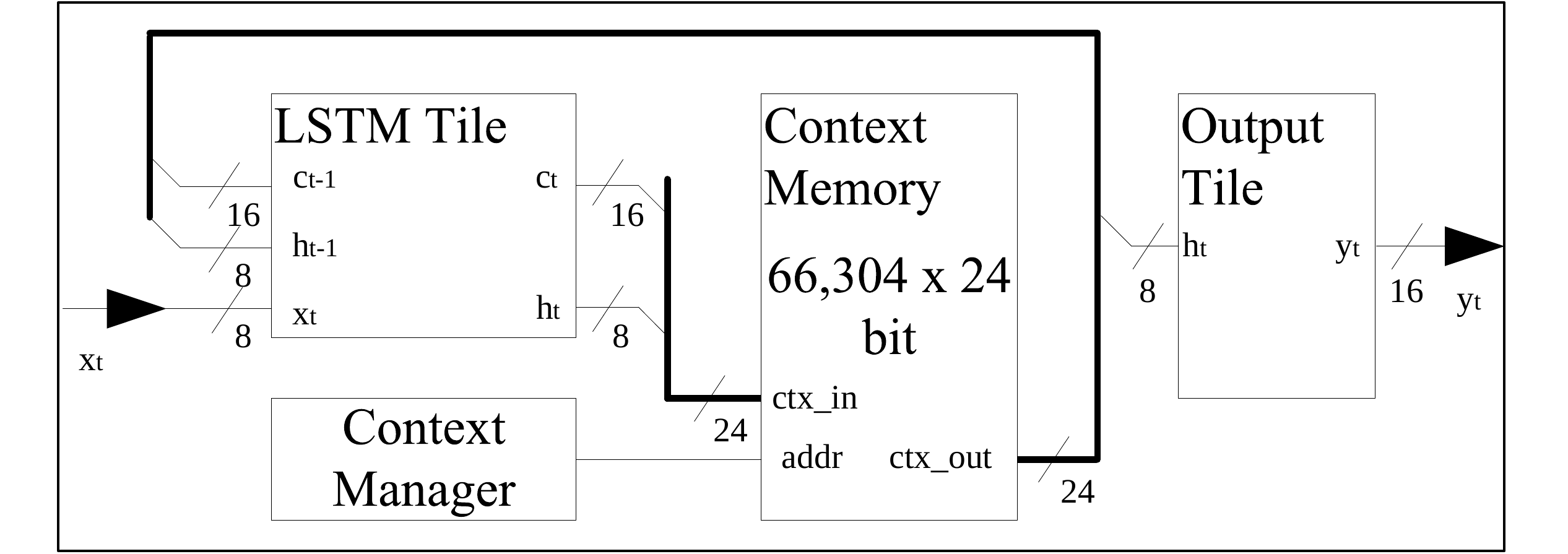}}
\vskip -0.08in
\caption{Hardware architecture for implementing RNNs.}
\label{fig_lstmStructrue}
\end{center}
\vskip -0.4in
\end{figure}

The proposed algorithm is implemented on a Xilinx ZC706 evaluation board that equips an XC7Z045 FPGA. The FPGA embeds an ARM CPU in addition to configurable logic circuits. \figurename~\ref{fig_lstmStructrue} shows the hardware architecture for implementing RNNs. Although the SR algorithm employs two RNN algorithms, our FPGA design implements only one LSTM tile and one output tile, which operate intermittently when the control signal is given.  Note that the RNN operation for the acoustic model is needed only once for each input speech frame whose length is normally 10-ms, but the character-level LM operates much more frequently to generate $N$-best hypotheses for different search paths. 

\subsection{Architecture and Algorithm}
\label{subsec_archAlg}

The standard LSTM with peephole connections is described in Algorithm \ref{alg_lstm}. The equations show that one LSTM RNN layer requires eight matrix-vector multiplications in each time step. 

\begin{figure}[t]
\vskip -0.08in
\begin{center}
\centerline{\includegraphics[width=\columnwidth]{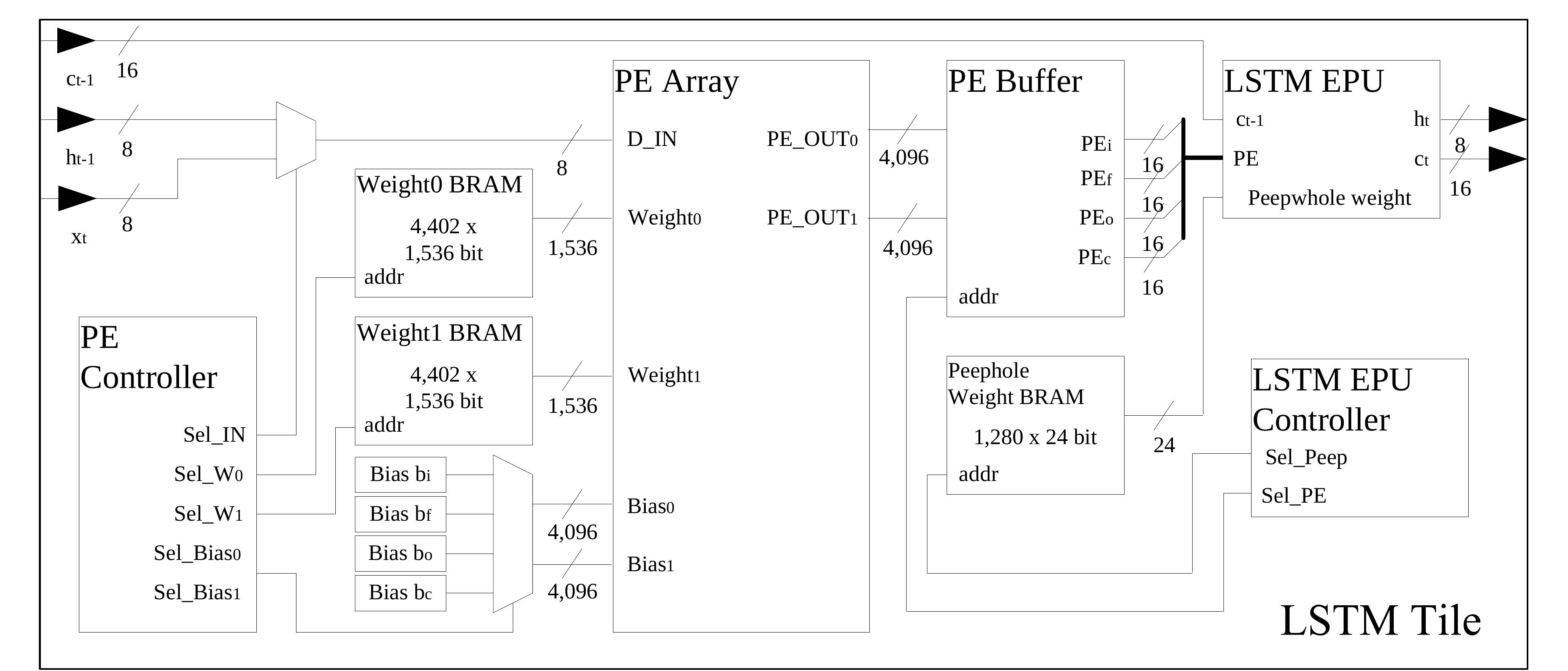}}
\vskip -0.08in
\caption{Structure of the LSTM tile.}
\vskip -0.08in
\label{fig_LSTMtile}
\end{center}
\vskip -0.08in
\end{figure}

The LSTM tile in \figurename~\ref{fig_LSTMtile} consists of two main processing modules; the processing element (PE) array calculates matrix-vector multiplications and the LSTM extra processing unit (LSTM EPU) conducts the rest of the calculations, such as applying element-wise products for peephole connections and evaluating activation functions. 

\begin{algorithm}[t]
\caption{LSTM equations with peephole connections: $\mathbf{x}$ is the input vector of the input layer, $\mathbf{h}$ is the output vector of the layer. The vector $\mathbf{i}$, $\mathbf{f}$ and $\mathbf{o}$ are activations of the input gate, forget gate and the output gate processed by the logistic sigmoid function $\sigma$, respectively. $\mathbf{c}$ represents the activation of the cell and $\widetilde{\bm{c}}_{t}$ is the candidate memory cell. The vector $\mathbf{b}$ stands for the bias. The subscript $t$ is the current data where $t-1$ denotes the data from the previous time step. $W$ is the model parameter matrix and $\widetilde{W}$ is the diagonal model parameter matrix. The operator $\odot$ is an element-wise multiplication, and $tanh$ is a hyperbolic tangent. }
\label{alg_lstm}
\begin{algorithmic}
\STATE{$\mathbf{i}_{t} = \sigma({W}_{xi}\,\mathbf{x}_{t}+{W}_{hi}\,\mathbf{h}_{t-1}+\widetilde{{W}}_{ci}\,{c}_{t-1}+\mathbf{b}_{i})$}
\STATE{$\mathbf{f}_{t} = \sigma({W}_{xf}\,\mathbf{x}_{t}+{W}_{hf}\,\mathbf{h}_{t-1}+\widetilde{{W}}_{cf}\,{c}_{t-1}+\mathbf{b}_{f})$}
\STATE{$\mathbf{o}_{t} = \sigma({W}_{xo}\,\mathbf{x}_{t}+{W}_{ho}\,\mathbf{h}_{t-1}+\widetilde{{W}}_{co}\,{c}_{t}+\mathbf{b}_{o})$}
\STATE{$\widetilde{\mathbf{c}_{t}} = tanh({W}_{xc}\,\mathbf{x}_{t}+{W}_{hc}\,\mathbf{h}_{t-1}+\mathbf{b}_{c})$}
\STATE{$\mathbf{c}_{t} = \mathbf{f}_{t}\odot\mathbf{c}_{t-1}+\mathbf{i}_{t}\odot\widetilde{\mathbf{c}}_{t}$}
\STATE{$\mathbf{h}_{t} = \mathbf{o}_{t}\odot tanh(\mathbf{c}_{t})$}
\end{algorithmic}
\end{algorithm}

\begin{figure}[t]
\begin{center}
\centerline{\includegraphics[width=\columnwidth]{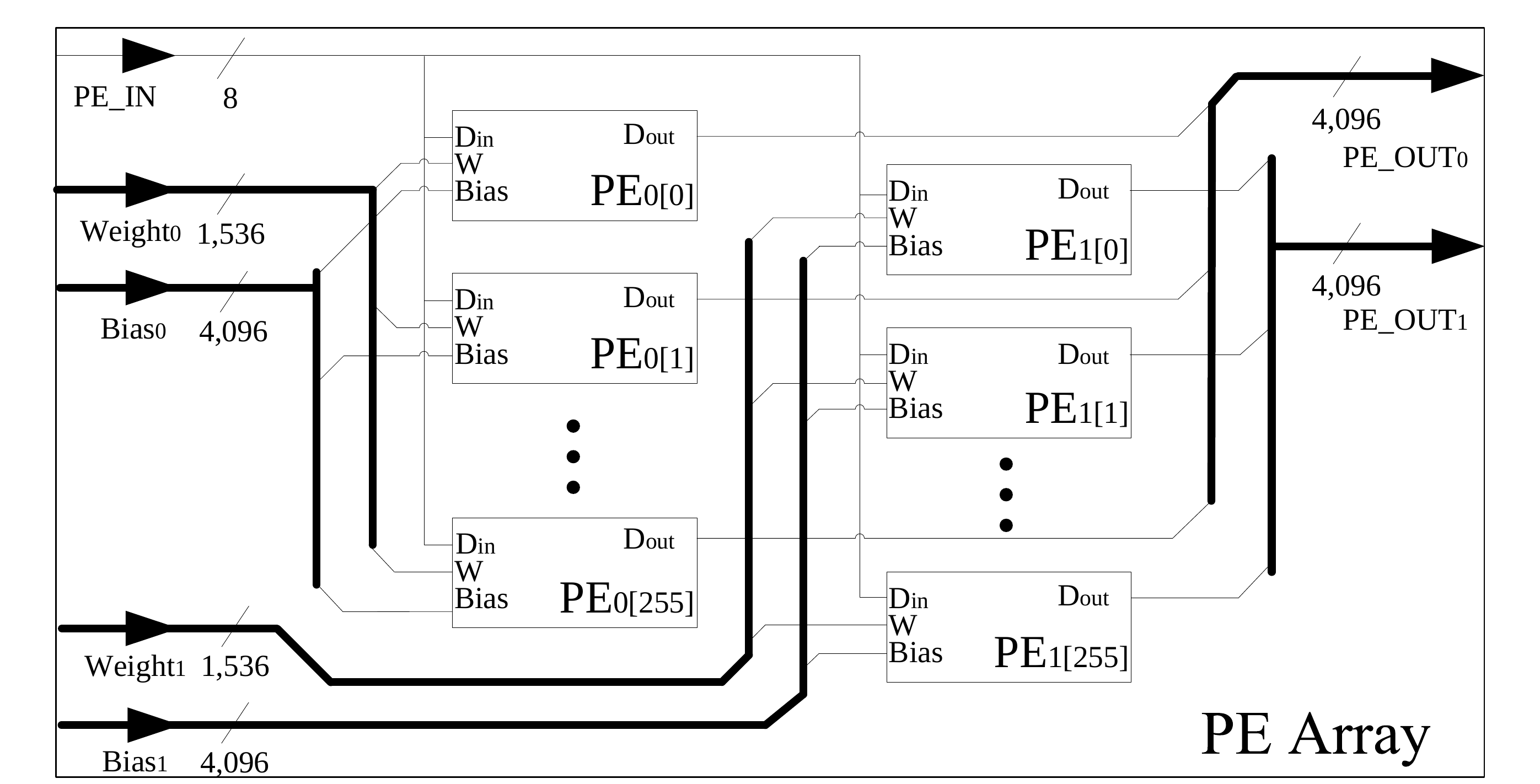}}
\vskip -0.08in
\caption{Structure of the processing element array.}
\label{fig_pearray}
\end{center}
\vskip -0.08in
\end{figure}

\begin{figure}[t]
\begin{center}
\centerline{\includegraphics[width=50mm]{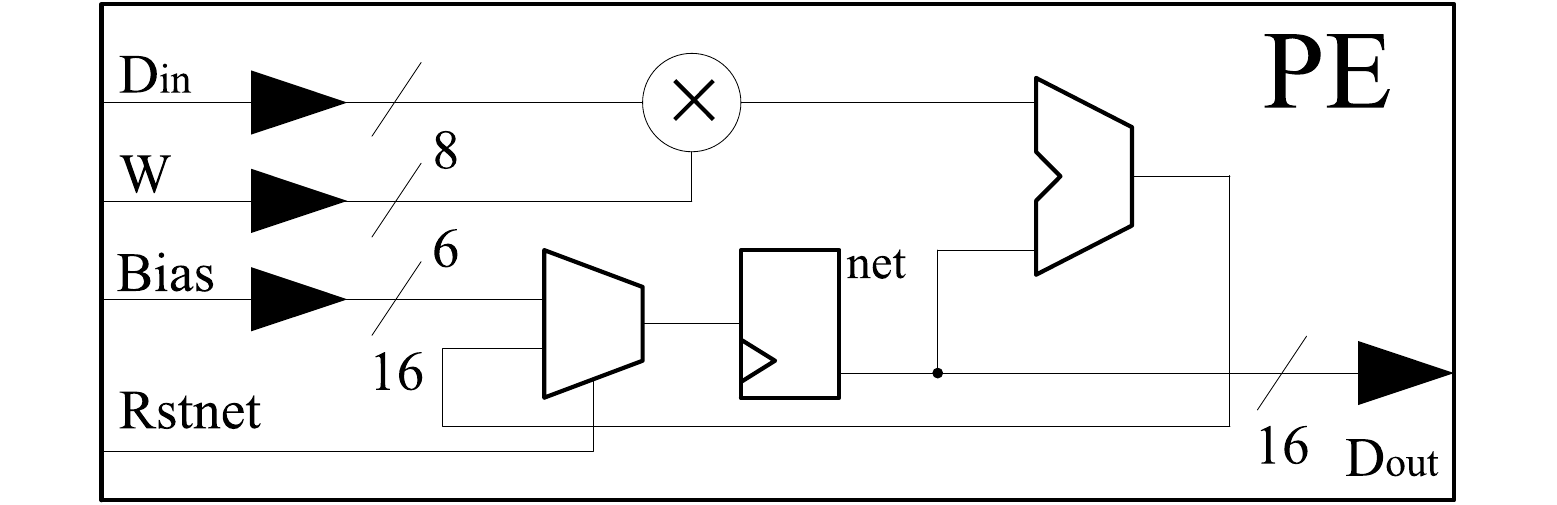}}
\vskip -0.08in
\caption{Structure of the processing element.}
\vskip -0.08in
\label{fig_PE}
\end{center}
\vskip -0.08in
\end{figure}

As shown in \figurename~\ref{fig_pearray}, the PE array consists of 512 PEs. The PE in \figurename~\ref{fig_PE} multiplies the input $D_{in}$ with the weight $W$ and adds the result with the partial sum stored in the accumulator where the bias values are preloaded \cite{park2016fpga}. The results of eight matrix-vector multiplications are stacked in the PE output buffer. We use four PE buffers, $PE_i$, $PE_f$, $PE_o$ and $PE_c$.

\begin{figure}[t]
\begin{center}
\centerline{\includegraphics[width=70mm]{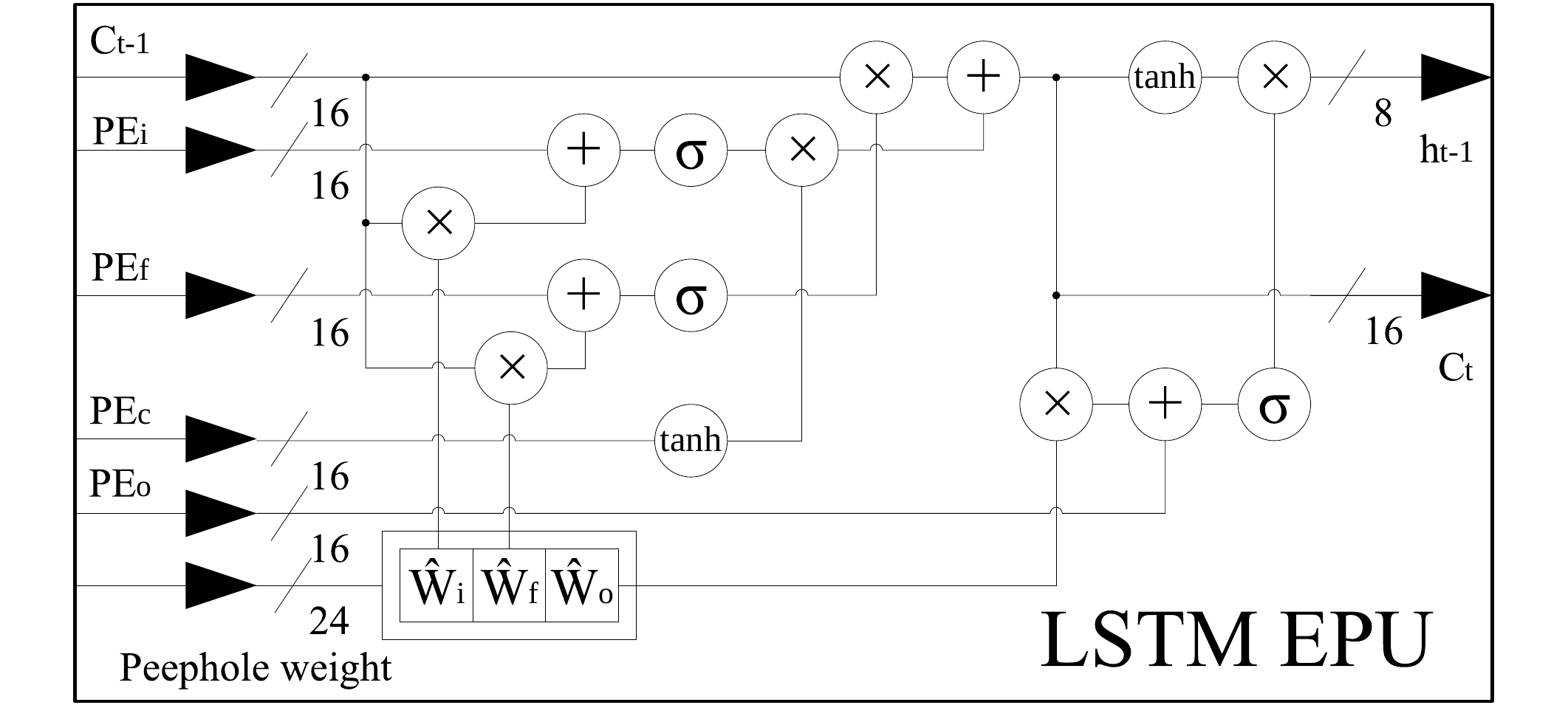}}
\vskip -0.08in
\caption{Structure of the LSTM extra processing unit.}
\label{fig_LSTMEPU}
\end{center}
\vskip -0.08in
\end{figure} 

The LSTM EPU shown in \figurename~\ref{fig_LSTMEPU} is implemented to manage the rest of the LSTM operations. The input ${c}_{t-1}$ represents the cell activation of the previous time step.

To implement the peephole connections in the LSTM, ${c}_{t-1}$ is multiplied with the peephole weights and added to $PE_i$ and $PE_f$ while ${c}_{t}$ is multiplied with the weights and added to $PE_o$.  Since the matrix-vector multiplication results are already stored in the PE buffers, the LSTM EPU and the PE array can operate independently. The activation functions in the LSTM EPU are implemented using lookup tables. In the proposed system, only one LSTM EPU is used because one output data is transmitted in each clock and all the operations in the LSTM EPU are element-wise ones. 

The output vector of the LSTM EPU is stored in the context memory. The stored contexts are used in the following operations and the beam search decoding. The number of stored contexts is the same as that of hypotheses in the beam search. 

The output tile is a fully connected layer that employs the same structure in \cite{park2016fpga}. The input of the output tile is the data stored in the context memory.

\subsection{Throughput of the LSTM tile}
\label{subsec_dataPath}

As shown in  \figurename~\ref{fig_pearray}, there are two PE arrays in the PE array block. Since there are eight matrix-vector multiplications, one RNN layer demands four matrix-vector multiplication cycles. Each PE array has 256 PEs and conducts a matrix-vector multiplication using the outer product method. 
The processing time of the LSTM depends on the dimension of the input vector because the outer product method supplies one input element at each clock. The input size of the first level RNN AM is 123 and that of the next layers is 256. Thus, the first layer processing of the RNN AM requires 246 $(=123\times4\div2)$ and 512 $(=256\times4\div2)$ clock cycles to conduct matrix-vector multiplications related with $\mathbf{x}_{t}$ and $\mathbf{h}_{t-1}$. The number of clock cycles for the next layer is 1,024. Note that there exists a small overhead to synchronize the system. The number of required clock cycles to process the RNN AM with three LSTM layers is 2,806 $(=758+1,024+1,024)$ and that of the RNN LM containing two LSTM layers is 1,596 $(=572+1,024)$, respectively.

\section{Experimental Results}
\label{sec_exp}

\subsection{Recognition Performance}
\label{subsec_expRecog}

To train the RNN AM, we use the standard WSJ \texttt{SI-284} training set. The utterances with verbalized punctuations are removed and odd transcriptions are filtered out. The final size of the training set is roughly 71 hours. For evaluation, the WSJ \texttt{eval92} (Nov'92 20k evaluation set) is used. The utterances in the evaluation set are sequentially concatenated to generate a single 42-minute input speech stream.  

The RNN AM is trained using the stochastic gradient descent (SGD) with 8 parallel input streams on a GPU \cite{hwang2015single}. 

The RNN AM uses a 40-dimensional log mel frequency filterbank feature with energy and their delta and double-delta, resulting in a 123-dimensional vector. The feature vector is computed every 10 ms over a 25 ms Hamming window and element-wisely normalized based on the statistics obtained from the training set. A centered sliding-window with 300-frame size is used to reduce the amplitude distortion effect from silence intervals. The RNN AM outputs a 31-dimensional vector representing  the probabilities of 26 upper case alphabet characters, 3 special characters for punctuation marks, the “end of sentence” symbol, and the CTC blank label. 

The RNN LM is trained with a text stream generated by concatenating randomly selected sentences in the WSJ non-verbalized punctuation text corpus where the EOS label is inserted between the sentences. The RNN LM is trained with AdaDelta \cite{zeiler2012adadelta} based SGD. The RNN LM uses a 30-dimensional vector where the current character-label is one-hot encoded and outputs a 30-dimensional vector which represents the probabilities of the following character-labels. 

The statistical tri-gram LM is generated with the IRSTLM \cite{federico2008irstlm} toolkit included in the KALDI speech recognition tool \cite{povey2011kaldi}. \texttt{build-lm.sh} and \texttt{compile-lm} in IRSTLM toolkit is used to generate a standard advanced research project agency (ARPA) file while applying the improved Kneser-Ney method \cite{kneser1995improved} for higher performance. We use the WSJ non-verbalized punctuation text corpus that contains 165 K words to build the LM. The generated 578-MB ARPA file is stored in the off-chip DRAM.

\begin{table}[t]
\caption{The WER and the CER performance (\%) of the SR algorithm with respect to the beam width.}
\vskip -0.08in
\label{tbl_optimize}
\begin{center}
\begin{small}
\begin{sc}
\begin{tabular}{c|c|c|c}
\hline
\renewcommand{\multirowsetup}{\centering}
\linebreakcell{Network}	&\linebreakcell{Word LM}	&	\linebreakcell{Beam width}	&	\linebreakcell{WER / CER }\\
\hline
\hline
\multirow{4}{*}{\linebreakcell{Small\\model}}	&	\multirow{2}{*}{none}	&	{128}	&	13.45 / 5.43	\\\cline{3-4}
	&	&	256	&	12.78 / 5.24	\\\cline{2-4}
	&	\multirow{2}{*}{applied}	&	{128}	&	12.65 / 5.36	\\\cline{3-4}
	& 	&	256	&	12.17 / 5.11	\\
\hline

\multirow{4}{*}{\linebreakcell{Large\\model}}	&	\multirow{2}{*}{none}	&	{128}	&	9.21/4.10	\\\cline{3-4}
	&	&	256	&	8.86 / 3.71	\\\cline{2-4}
	&	\multirow{2}{*}{applied}	&	{128}	&	9.07/4.05	\\\cline{3-4}
	& 	&	256	&	8.79 / 3.90	\\
\hline
\end{tabular}
\end{sc}
\end{small}
\end{center}
\vskip -0.08in
\end{table}
\begin{table}[t]
\caption{The WER and the CER performance (\%) with respect to the weight precision of the small-model when the beam is 128.}
\vskip -0.08in
\label{tbl_fixed}
\begin{center}
\begin{small}
\begin{sc}
\begin{tabular}{c|c|c}
\hline
\renewcommand{\multirowsetup}{\centering}
\linebreakcell{Word LM}	&	\linebreakcell{Weight precision}	&	\linebreakcell{WER / CER}\\
\hline
\hline
\multirow{4}{*}{none}	&	{floating}	&	13.45 / 5.43	\\\cline{2-3}
	&	fixed (6-bit)	&	15.06 / 5.97	\\\cline{2-3}
	&	fixed (5-bit)	&	16.13 / 6.50	\\\cline{2-3}
	&	fixed (4-bit)	&	20.18 / 8.03	\\\cline{2-3}
\hline
\multirow{4}{*}{applied}	&	{floating}	&	12.65 / 5.36	\\\cline{2-3}
	&	fixed (6-bit)	&	14.02 / 6.02	\\\cline{2-3}
 	&	fixed (5-bit)	&	15.20 / 6.47	\\\cline{2-3}
 	&	fixed (4-bit)	&	18.50 / 7.71	\\
\hline
\end{tabular}
\end{sc}
\end{small}
\end{center}
\vskip -0.08in
\end{table}
The word error rate (WER) and character error rate (CER) performances of the proposed system with respect to the size of the RNNs and the beam width are shown in \tablename~\ref{tbl_optimize}. The small-model represents the system with 3$\times$256 RNN AM and  2$\times$256 RNN LM while the large-model employs 4$\times$512 RNN AM and 2$\times$512 RNN LM. The table shows that the performance improves when the beam width or the network size increases. Also, combining the word-level LM improves the performance especially when the network size is small.

The best floating-point performance of our algorithm in \tablename~\ref{tbl_optimize} shows the WER of 8.79 \% which is higher than
the state of the art result, 7.34 \% \cite{miao2015eesen}, but ours supports delay free real-time SR. 
Of course, the best advantage we expect is the energy efficiency since we do not employ a WFST network which demands a large amount of computation and memory accesses. Note that the algorithm in \cite{miao2015eesen} is not for real-time speech recognition task, and  employs a bidirectional structure that shows better performance over the unidirectional structure. The algorithm also uses the WFST decoding network to combine the results of acoustic modeling, lexicon, and the word-level LM. Note that the compared system does not use the character-level RNN because the WFST network embeds the lexicon. However, the WFST-based decoding demands a large memory space to search, and thus the algorithm is hard to be power efficient. On the other hand, our algorithm employs the character-level LM in addition to the word-level LM, and uses simple beam-search in decoding that requires far less memory. The RNNs of the proposed algorithm are implemented using only on-chip memory for energy efficiency. Note that the recognition performance of our system can be further improved by employing larger RNNs or increasing the beam width.

The SR algorithm is implemented on an XC7Z045 FPGA that has 2.18 MB on-chip memory. In our experiment, the number of parameters for the small-model is 2.3 M while that of the large-model is 15.1 M. The retraining based fixed-point optimization is applied to reduce the precision of weights. \tablename~\ref{tbl_fixed} shows the performance of the systems that employ fixed-point weights, where the precision of the signal and the LSTM cells are fixed to 8 and 16 bits, respectively. The table shows that rescoring with the word-level LM is also effective for the systems that employ fixed-point weights. The FPGA can only accommodate up to 6-bit weights, which demands only 1/5 of the memory space required for floating-point implementations with about 1.5\% WER increase. The size of the parameters with 6-bit precision is about 1.1 MB, which can be stored in the on-chip memory of Xilinx XC7Z045.

\subsection{FPGA Implementation Performance}
\label{subsec_expEff}

The FPGA implements the small-model with the beam width of 128. Note that the large-model based system can be implemented using an ultra-scale FPGA \cite{mehta2013xilinx}. In our implementation, the programmable hardware operates at 100-MHz and the CPU runs with a 800-MHz clock to conduct the $N$-best search. The FPGA resource utilization result is shown in \tablename~\ref{tbl_HWutilization}.

The implemented system requires one RNN AM operation for each 10 ms speech frame (100 times per second). However, the RNN  for character-level LM is needed only when character transition occurs, whose frequency is usually no more than 30 times per second in our experiments. Assuming 128 beams, this translates about 3,840 RNN LM operations per second. Thus, the number of clock cycles for achieving a real time with conservative estimation is about 6.4 M $(=100\times2,806+3,840\times1,596)$ per second. 
Note that silence period does not generate any transition, thus no RNN LM is demanded.

\begin{table}[t]
\caption{FPGA resource utilization of implemented SR system.}
\vskip -0.08in
\label{tbl_HWutilization}
\begin{center}
\begin{small}
\begin{sc}
\begin{tabular}{c|c|c|c|c}
\hline
\multirow{2}{*}{\linebreakcell{Network\\/model}}	&	\multicolumn{4}{c}{Resource} \\\cline{2-5}
	&	{FF}	&	{LUT}	&	{BRAM}	&	{DSP}\\
\hline
\hline
{\linebreakcell{Small\\model}}	&	{88,947}	&	{134,031}	&	{510}	&	{512}\\
\hline
{\linebreakcell{Large\\model}}	&	{148,453}	&	{227,185}	&	{2,001}	&	{1,567}\\
\hline
\hline
{\linebreakcell{Xilinx\\XC7Z045}}	&	{437,200}	& {218,600}	& {545}	& {900}\\
\hline

\end{tabular}
\end{sc}
\end{small}
\end{center}
\vskip -0.08in
\end{table}
\begin{table}[t]
\caption{Power consumption ($W$) of implemented SR system.}
\vskip -0.08in
\label{tbl_powerConsumption}
\begin{center}
\begin{small}
\begin{sc}
\begin{tabular}{c|c|c}
\hline
\multirow{2}{*}{\linebreakcell{Usage}}	&	\multicolumn{2}{c}{Network} \\\cline{2-3}
	&	{\linebreakcell{Small-model}}	&	{\linebreakcell{Large-model}}	\\
\hline
\hline
{Clocks}	&	{0.375}		&	{0.574}\\
\hline
{Signals}	&	{1.166}		&	{4.175}\\
\hline
{Logic}	&	{0.677}		&	{1.099}\\
\hline
{BRAM}	&	{0.836}		&	{2.386}\\
\hline
{DSP}	&	{0.574}		&	{1.409}\\
\hline
{PS7}	&	{1.636}		&	{-}\\
\hline

{device static}	&	{0.306}		&	{2.194}\\
\hline
\hline
{Total}	&	{5.570}	&	{11.837}\\
\hline
\end{tabular}
\end{sc}
\end{small}
\end{center}
\vskip -0.08in
\end{table}
\tablename~\ref{tbl_powerConsumption} shows the power consumption measured by the Xilinx simulation tool. The actual power consumption of the small-model based SR measured on the evaluation board is 9.24 $W$ including that in the DRAM and peripherals, while achieving $\times$ 4.12 real-time speed. Our implementation consumes some extra cycles for communication.

We compare our FPGA implementation with that of a high-end GPU, NVIDIA GeForce Titan X. In the GPU based implementation, the time to evaluate the 42-minute WSJ \texttt{eval92} evaluation set is 12.5 minute, which means $\times$3.36 real-time speed, while utilizing about 30 \% of GPU resource. Note that the throughput of the GPU can be increased by processing multiple input speech utterances. However, our FPGA based system shows better recognition speed by efficiently utilizing hardware resources even when processing a single speech stream. The power consumption of the GPU based system is about 80 $W$ which is much higher than ours.

\section{Concluding Remarks}
\label{sec_conclusion}

In this paper, an RNN based real-time speech recognition system is implemented on an FPGA. The algorithm employs the RNNs for acoustic modeling and character-level language modeling, and is optimized for real-time operations using unidirectional RNNs. The vocabulary size of the speech recognition is unlimited since the character-level RNN can dictate out of vocabulary words. A statistical word-level language model is also employed to improve the recognition performance. The models are integrated using a simple tree-based search algorithm without employing a hidden Markov model or weighted finite state transducers. The weights of the RNNs are quantized to 6 bits. The RNNs are implemented using an array of processing elements for high throughput matrix-vector multiplications. The RNNs implemented on the FPGA only use on-chip memory. The implemented speech recognition system on Xilinx XC7Z045 can achieve approximately 4.12 times of the real-time speed when 100 MHz clock is used while consuming only 9.24 $W$ of power. When compared to a high-end GPU based system, the power efficiency is considered about 10 times higher.

\section*{Acknowledgment}

This work was supported in part by the Brain Korea 21 Plus Project and the National Research Foundation of Korea (NRF) grant funded by the Korea government (MSIP) (No. 2015R1A2A1A10056051).



\bibliographystyle{IEEEtran}
\bibliography{reference}
%
%
%

\end{document}